\documentclass[11pt]{article}

\usepackage[most]{tcolorbox}
\usepackage{amsmath,amssymb}
\usepackage{xcolor}
\usepackage[preprint]{acl}

\usepackage{times}
\usepackage{latexsym}

\usepackage[T1]{fontenc}

\usepackage[utf8]{inputenc}

\usepackage{microtype}

\usepackage{inconsolata}

\usepackage{graphicx}

\usepackage{subcaption}

\newtcolorbox{sampleproblembox}{
  enhanced,
  breakable,
  width=\columnwidth,
  colback=blue!3,
  colframe=blue!45!black,
  boxrule=0.5pt,
  arc=1.2mm,
  left=1.2mm,right=1.2mm,top=0.8mm,bottom=0.8mm,
  title=Sample Problem,
  fonttitle=\bfseries,
  before upper=\small
}

\newtcolorbox{samplesolutionbox}{
  enhanced,
  breakable,
  width=\columnwidth,
  colback=gray!4,
  colframe=black!55,
  boxrule=0.5pt,
  arc=1.2mm,
  left=1.2mm,right=1.2mm,top=0.8mm,bottom=0.8mm,
  title=Sample Solution,
  fonttitle=\bfseries,
  before upper=\footnotesize
}

\newcommand{\solstep}[1]{\par\medskip\noindent\textcolor{blue!60!black}{\textbf{#1}}\par\smallskip}

\title{Too long; didn't solve}

\author{
\begin{tabular}{c@{\hspace{2.5em}}c@{\hspace{2.5em}}c}
\textbf{Lucía M. Cabrera\textsuperscript{1,2,$\dagger$}} &
\textbf{Isaac Saxton-Knight\textsuperscript{2,$\dagger$}} &
\textbf{Jocelyn D'Arcy\textsuperscript{2}} \\[0.6em]
\multicolumn{3}{c}{{\normalfont
\textsuperscript{1}Instituto Balseiro \quad \textsuperscript{2}Poindexter Labs}} \\[0.3em]
\multicolumn{3}{c}{{\normalfont
\textsuperscript{$\dagger$}Equal contributions}}
\end{tabular}
}

\begin{document}
\maketitle

\begin{abstract}

Mathematical benchmarks consisting of a range of mathematics problems are widely used to evaluate the reasoning abilities of large language models, yet little is known about how their structural properties influence model behaviour. In this work, we investigate two structural length variables, prompt length and solution length, and analyse how they relate to model performance on a newly constructed adversarial dataset of expert-authored mathematics problems. Across five evaluated models, we find that both prompt length and solution length are positively associated with model failure. These associations are statistically significant but modest, and we interpret them as descriptive rather than causal. We also include a secondary, exploratory analysis of cross-model disagreement. Because disagreement measures based on variance are mechanically constrained by mean failure, we treat this part of the analysis cautiously. Overall, our main finding is that structural length is linked to empirical difficulty in this benchmark, suggesting that length should be considered as a potential confounder when interpreting mathematical model evaluations.

\end{abstract}

\section{Introduction}

The modern landscape of large language model (LLM) evaluation is increasingly shaped by advances in reasoning-oriented models.
In the context of mathematical reasoning, benchmarks such as GSM8K \citep{gsm-8k}, MATH \citep{MATH}, MathArena \citep{math-arena}, OlympiadBench \citep{olympiadbench}, AGIEval \citep{agieval} and MathVista \citep{mathvista} have become standard tools for evaluating the capabilities of LLMs. These typically comprise problems designed to stress multi-step reasoning chains and often require a single numerical or symbolic answer. Sustained improvements in model performance have also recently motivated the development of more sophisticated benchmarks, such as FrontierMath \citep{frontiermath}, BIG-Bench Extra Hard \citep{bbeh} and GSM-Symbolic \citep{gsm-symbolic}, aimed at probing the limits of current systems across a range of metrics.

Evaluation results on these benchmarks are typically reported by aggregating performance across categorical variables such as topic or difficulty level. While informative, these labels are discrete, coarse, and partly subjective, which may obscure item-level structural patterns in model performance \citep{gsm-infty}. Related work has also highlighted broader challenges in evaluating reasoning systems, including issues of verification, reliability, and static benchmark design \citep{proof-or-bluff,dynabench}. These concerns point to a broader need for evaluation analyses that look beyond headline accuracy and examine how properties of benchmark items themselves shape model behaviour.

In this work, we focus on one such class of item-level properties: continuous structural features of problems. Specifically, we study the word count of the problem statement and of the associated human-authored solution. The idea of studying prompt-level features appears extensively in the literature \citep{lost-middle, prosa, state-of-what-art, ruler, infbench}, but has rarely been exploited in the specific arena of LLM mathematical reasoning. Unlike categorical labels, length-based features are simple, reproducible, and model-agnostic measurements. They do not replace semantic notions such as topic or difficulty, but can reveal systematic variation hidden by coarser labels.

We analyse how these structural variables relate to model failure on an adversarially constructed dataset of original expert-authored mathematics problems, and, secondarily, how they relate to cross-model disagreement. This is relevant for the broader evaluation community because apparent differences in mathematical ability may partly reflect structural properties of benchmark items rather than mathematical reasoning alone. As such, these item-level properties may be unreported confounders that distort current evaluations and leaderboards \citep{leaderboard-illusion}.

\section{Our dataset}

Our dataset comprises a collection of 607 complex mathematics problem--solution pairs, crafted by a team of domain experts, including mathematics researchers and IMO medalists, between November 2025 and January 2026, and specifically designed to induce failures in state-of-the-art large language models. To qualify for this benchmark, problems were required to have a single integer-valued final answer and a minimum of three reasoning steps. During the pre-screening stage, candidate problems were tested against Gemini 2.5 Pro and GPT 5.2 Thinking, and those that did not induce failures were rejected or revised. Once this stage was cleared, items now comprising a problem plus a full step-by-step solution, both developed by the same author, went through a dual-review process: they were first screened for ambiguity by an LLM agent, and later revised and accepted or sent back for revision by an expert. Finally, an LLM-based search agent screened candidate problems for plagiarism and overlap with public sources, and its findings were verified by a human reviewer. None of the problems tested in this work were drawn from publicly available sources, nor were they accessible online at the time of evaluation, safeguarding the analysis from data contamination. Because the benchmark consists of original olympiad-style and IMO-flavoured problems written specifically for this evaluation, the structural effects we observe are unlikely to be artifacts of repeated exposure to familiar public-domain items. A sample of the style of problem-solution pair used in this evaluation can be found in Appendix~\ref{ap:samples}.  

Each item in this collection is categorised by a topic: Geometry, Combinatorics and Discrete Mathematics, Counting and Probability, Algebra, Linear Algebra, Number Theory and Calculus. Prompts are in turn also labelled by the problem-writer as either high school, undergraduate or graduate level. Regardless of their assigned level tag, all problems require complex multi-step reasoning, and were deliberately written across a range of topics and grade-levels in order to diversify the dataset. 

The evaluation pipeline leveraged five different models: GPT 5, GPT 4.1, GPT OSS 120B, Gemini 2.5 Flash, and Claude Sonnet 4.5. Each model performs five independent attempts per task, and we store this information as a fail count $k_{i,m}$ in $\lbrace0,1,2,3,4,5 \rbrace$ for problem $i$ and model $m$.

We selected these five models to balance coverage of strong proprietary systems with an optimized open-weights model, while keeping the repeated-attempt evaluation computationally feasible. Since each model was run five times on each task, the evaluation required a substantially larger number of model calls than a single-pass benchmark. We therefore prioritised depth of evaluation per model over a broader but shallower comparison across many systems. The resulting cross-model analyses should be interpreted as specific to this model set rather than as universal statements about all LLM architectures.

It is worth emphasizing that problems in this dataset were designed to have a single, ground-truth final answer. Thus, each run returns a binary fail/success result depending on whether the final answer was reached.

We define the failure fraction per problem, per model, as the average fail count,
\begin{equation}\label{eq: fail_frac_x_im}
x_{i,m} = \frac{k_{i,m}}{5}\, ,    
\end{equation}
so that $x_{i,m} \in \{0,0.2,0.4,0.6,0.8,1.0\}$. This quantity captures empirical instability and error rate of model $m$ on problem $i$. To summarise the empirical difficulty of problem $i$, we define its mean failure fraction across models,
\begin{equation} \label{eq: mean_failure_mu_i}
    \mu_i = \frac{1}{M} \sum_{m=1}^{M} x_{i,m}\, ,
\end{equation}
where $M$ is the total number of models. For our dataset, $M=5$. We include a summary of model performance on this dataset, collected as mean failure fraction per model $\mu_{m} = \frac{1}{607} \sum_{i=1}^{607} x_{i,m}$, in Table \ref{tab: model_mean_failures}.
\begin{table}[htb!]
\centering
\small
\begin{tabular}{lc}
\hline
\textbf{Model} & \textbf{Mean failure fraction $\mu_{m}$} \\
\hline
GPT 5 & 0.416 \\
GPT 4.1 & 0.860 \\
GPT OSS 120B & 0.690 \\
Gemini 2.5 Flash & 0.701 \\
Claude Sonnet 4.5 & 0.680 \\
\hline
\end{tabular}
\caption{Mean failure fraction across the 607 evaluated problems for each model.}
\label{tab: model_mean_failures}
\end{table}

\begin{figure}[htb!]
  \includegraphics[width=\columnwidth]{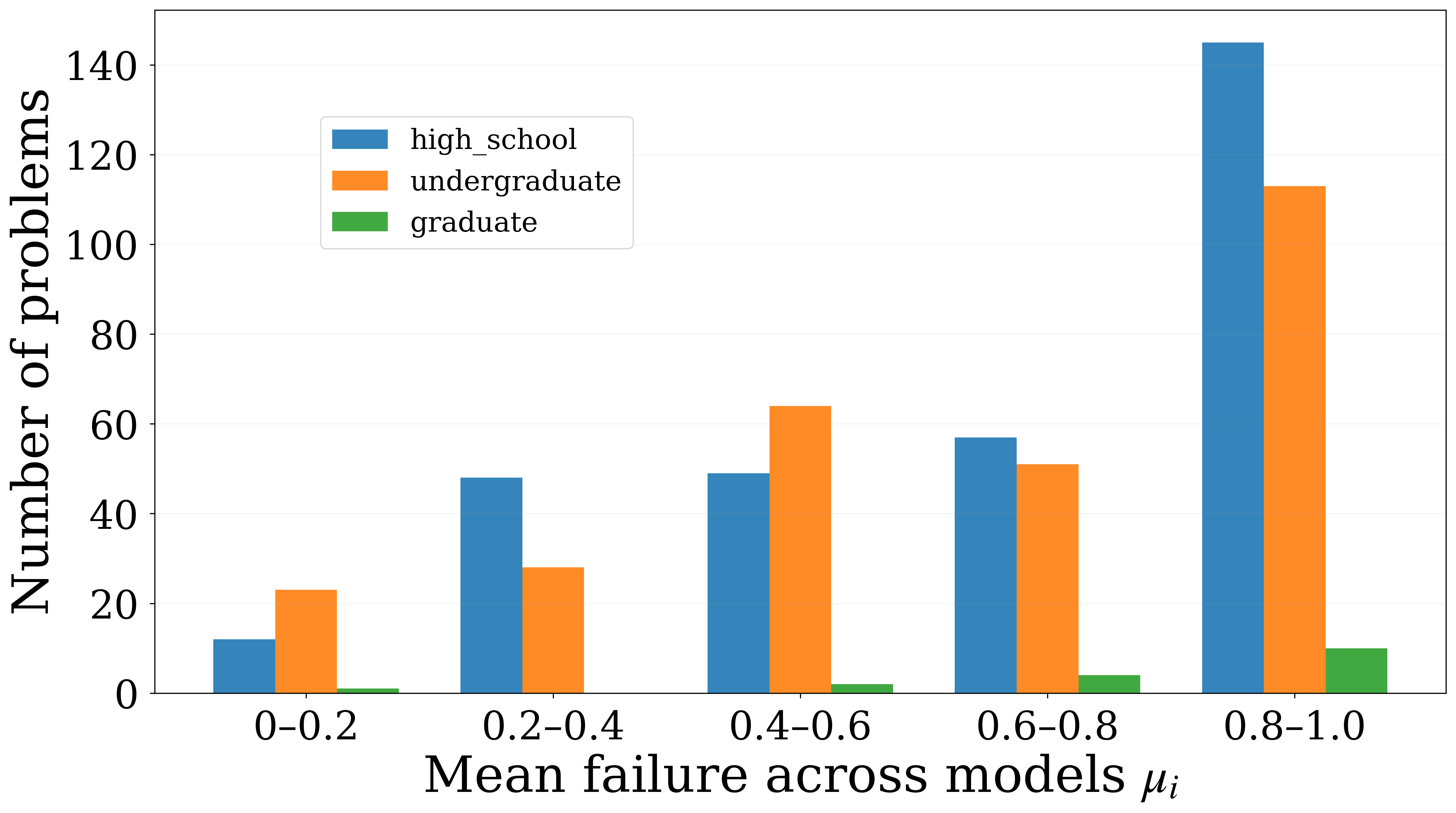}
  \caption{Number of problems, color-coded by level, producing mean failure rate $\mu_{i}$ across all five models.}
  \label{fig:problems_vs_meanfail_All}
\end{figure}

Figure~\ref{fig:problems_vs_meanfail_All} contains a visual of the distribution of the total number of problems in our dataset across their given level labels and mean failure fraction.
As is evident from Figure \ref{fig:problems_vs_meanfail_All}, the dataset contains a large number of high-school- and undergraduate-labelled problems, and the histogram is skewed toward the high mean-failure region. This reflects the adversarial nature of the collection: tasks were deliberately authored by human experts to be difficult for current models to solve. In other words, the benchmark is not merely a passive aggregation of existing problems, but a purpose-built evaluation set designed to expose weaknesses in contemporary mathematical reasoning systems.

Statistics on datasets of this kind are often performed based on somewhat arbitrarily given tags, such as the aforementioned topic and grade level. The labelling of a given problem under a certain grade level is largely subjective and dependent on several factors, such as differences in education systems. Similarly, topic labels are very coarse variables. Problems that mix different topics are reduced to a single, again subjective choice among all possibilities, and information is thus lost. All in all, the discrete nature of these problem-level variables makes for but a limited analysis.

Throughout this work, instead, we choose to focus on two objectively measurable quantities: the word count in the problem statement and in its given solution. In the following sections, we study the impact of these structural variables on model performance, with a secondary analysis of cross-model disagreement.

\section{Structural length as a correlate of difficulty}

\begin{figure*}[t]
\centering

\begin{subfigure}{0.48\linewidth}
\centering
\includegraphics[width=\linewidth]{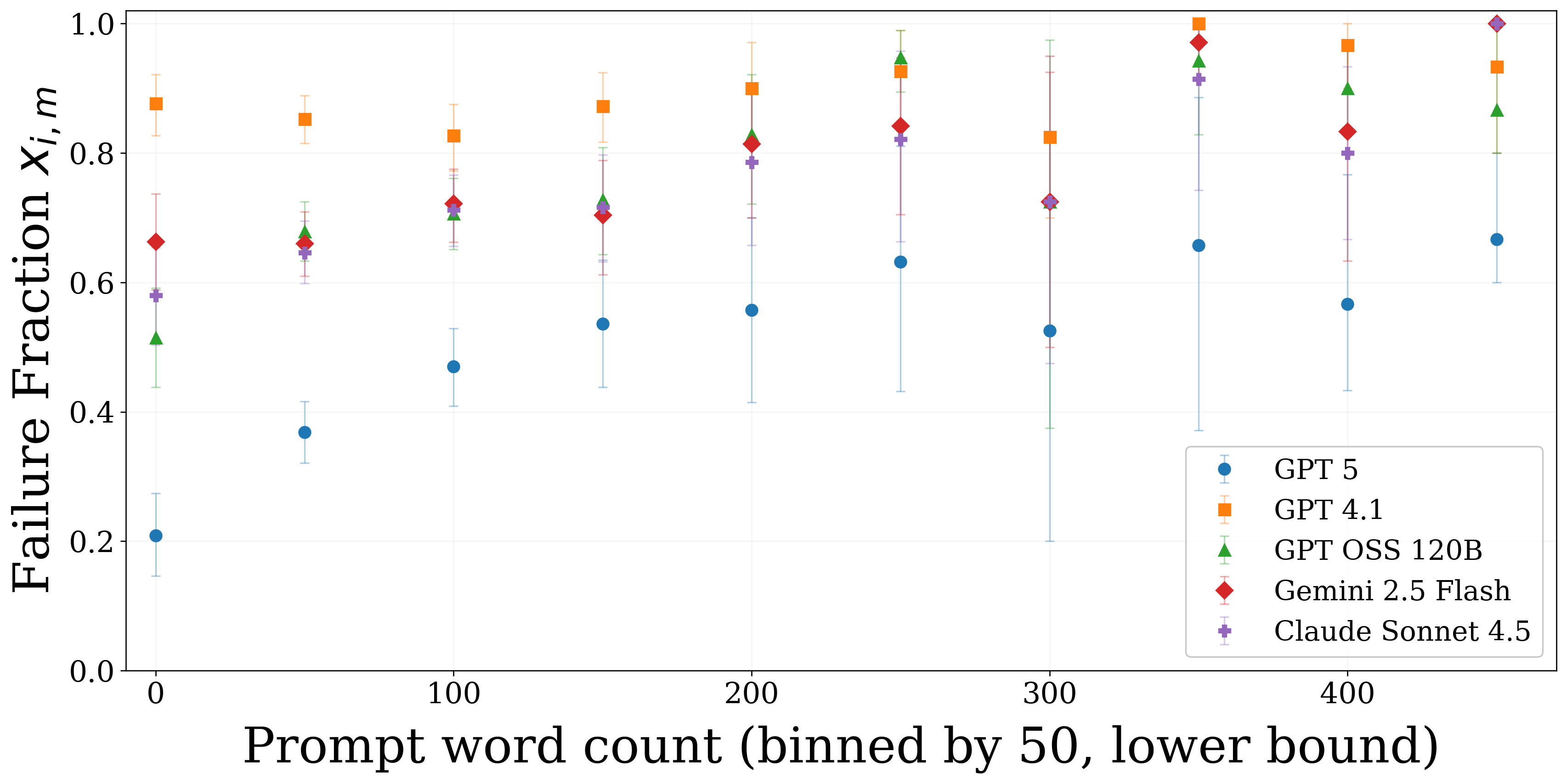}
\caption{Prompt length vs. failure fraction.}
\label{fig:x_vs_promptlength}
\end{subfigure}
\hfill
\begin{subfigure}{0.48\linewidth}
\centering
\includegraphics[width=\linewidth]{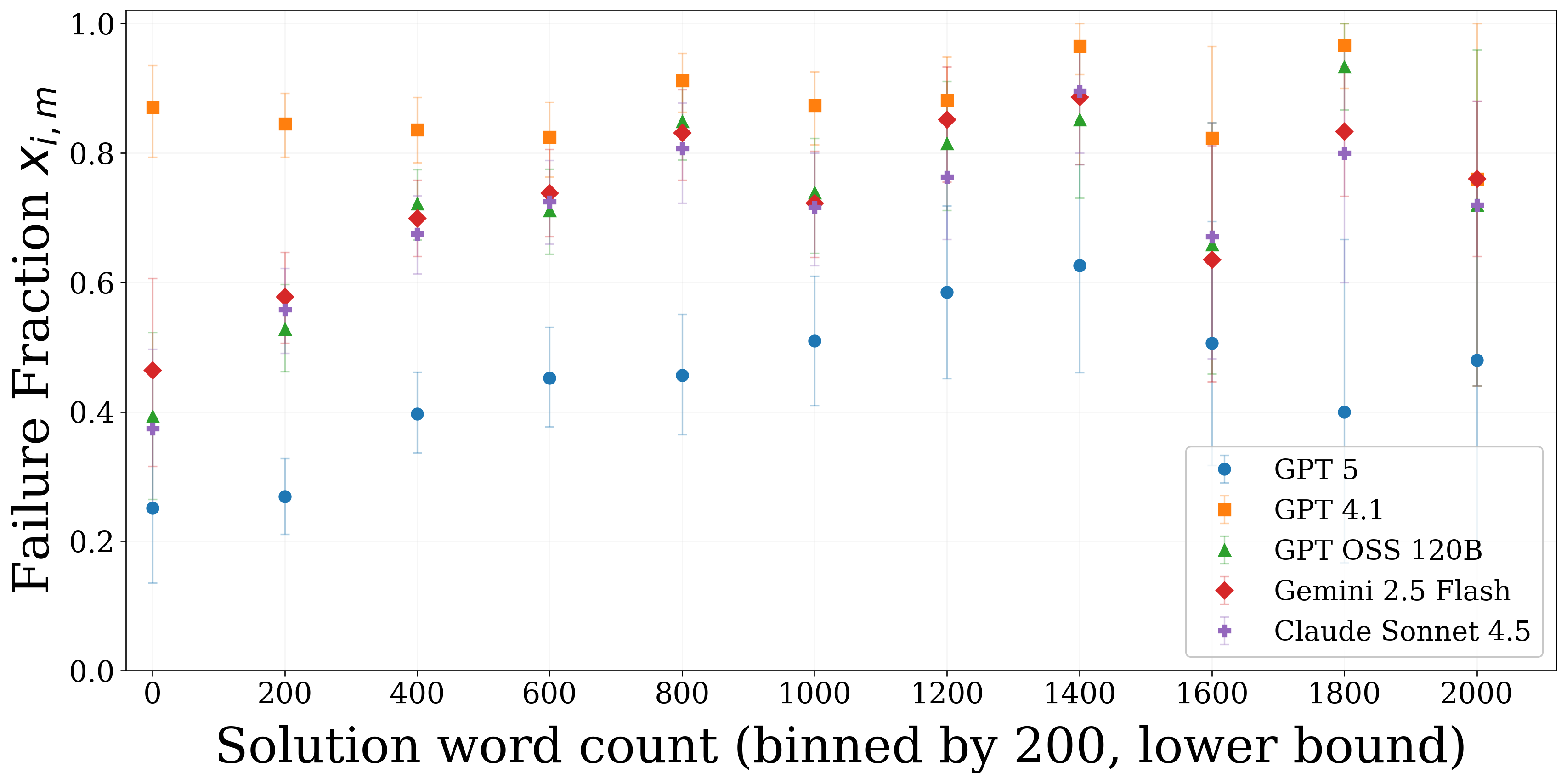}
\caption{Solution length vs. failure fraction.}
\label{fig:x_vs_sollength}
\end{subfigure}

\caption{Failure fraction $x_{i,m}$ as a function of structural length variables. Error bars show $95\%$ bootstrap confidence intervals.}
\label{fig:x_vs_length_prompt_sol}

\end{figure*}

A preliminary analysis reveals a visual association between model failure fraction $x_{i,m}$ and two length-based quantities: the word count of the problem statement and the word count of its step-by-step reference solution. As shown in the binned plots in Figures~\ref{fig:x_vs_promptlength} and~\ref{fig:x_vs_sollength}, model performance generally degrades as these structural variables increase, although local bin-level fluctuations remain visible, especially in sparsely populated length ranges. The data become particularly sparse in the long-prompt tail, which explains the wider error bars in that region.

This overall trend is visible across all models analysed, despite baseline differences in average ability. Although prompt lengths vary substantially across problems, all tasks remain comfortably within the context-window limits of the evaluated models, so the observed degradation cannot be attributed simply to exceeding model input capacity. The natural question is therefore whether prompt and solution length capture structural aspects of empirical problem difficulty, or whether they primarily act as proxies for latent mathematical complexity.

A secondary question is whether these structural variables are also related to cross-model disagreement, although such analyses require care because disagreement measures are mechanically constrained by mean failure.

In what follows, we analyse prompt and solution lengths as structural correlates of empirical difficulty, and then provide a more tentative, exploratory analysis of their relationship to cross-model disagreement.

\subsubsection*{Prompt Length}

To validate our observations, we analyse the relationship between prompt length, measured as the raw word count in problem $i$, and mean failure per problem, as defined in \eqref{eq: mean_failure_mu_i} across all models. Spearman’s rank correlation yields $\rho(\text{prompt length}, \mu) = 0.28$, with $p \ll 0.001$, which indicates a small but statistically significant positive association. Longer prompts are therefore more likely to produce model errors on average, which confirms the visual trend in Figure \ref{fig:x_vs_promptlength}.

This positive association is observed across all model families, suggesting that sensitivity to verbosity is not exclusive to a single architecture.

\subsubsection*{Solution Length}

A similar relationship emerges when considering the length of the provided solution. Spearman's rank correlation between solution length and mean failure is $\rho(\text{solution length}, \mu) = 0.32$, with $p \ll 0.001$, indicating a small-to-moderate positive association between the two. As with prompt length, longer solutions are positively correlated with higher model failure on average, consistent with the visual trend in Figure \ref{fig:x_vs_sollength}.

Solution length may more directly reflect underlying mathematical complexity than prompt length does. Since the model is not shown the reference solution, its length should be interpreted as a proxy for the amount of reasoning required to solve the problem, rather than as a direct input-side property of the task. 

An important limitation of this length variable is that it depends on the particular reference solution provided. Many mathematics problems admit multiple valid solution strategies, which may differ substantially in length while leading to the same final answer. Consequently, associations between solution length and difficulty should be interpreted cautiously, as the former may reflect authorial choices as well as intrinsic problem difficulty.

\subsection{Exploratory analysis of cross-model disagreement}

Having established that both prompt length and solution length are associated with model failure, we now briefly examine whether these structural features are also related to cross-model disagreement.

To quantify disagreement on a given problem, we define the variance of failure fractions across models,
\begin{equation} \label{eq: cross_model_var_i}
\mathrm{Var}_i = \frac{1}{M}\sum_{m=1}^{M}(x_{i,m} - \mu_{i})^{2} \, .
\end{equation}

This quantity summarises cross-model disagreement on problem $i$. A value near zero indicates universal behaviour across models, that is, all models either fail, succeed, or exhibit the same failure rate on the given problem, whereas higher values indicate stronger separation in performance. However, because each $x_{i,m}\in[0,1]$, the attainable magnitude of $\mathrm{Var}_i$ depends on $\mu_i$ and satisfies the bound
\[
\mathrm{Var}_i \le \mu_i(1-\mu_i) \, .
\]
Consequently, the cross-model variance should be interpreted as a disagreement measure whose feasible range depends on mean empirical difficulty.

Because of this mechanical mean--variance coupling, raw correlations between structural length and cross-model variance are difficult to interpret directly. In a dataset concentrated toward harder problems, any variable that is positively associated with mean failure will tend to exhibit a downward-biased raw correlation with variance. For this reason, we do not treat raw variance correlations as a central result.

As a more informative exploratory summary, we define a normalised variance score
\begin{equation} \label{eq: normalised_Var}
\widetilde{\mathrm{Var}}_i
=
\frac{\mathrm{Var}_i}{\mu_i(1-\mu_i)}\, .
\end{equation}
This metric measures the fraction of the theoretical maximum variance achieved by problem $i$ at its observed difficulty level. This quantity is only defined for tasks with $0<\mu_i<1$, so the normalised analysis excludes tasks that were empirically solved by all models or failed by all models. In our dataset, this leaves 517 tasks.

Under this normalisation, prompt length retains a weak negative association with difficulty-adjusted cross-model disagreement: $\rho(\text{prompt length},\widetilde{\mathrm{Var}})=-0.21$, $p \ll 0.001$. 

We interpret these results cautiously. The normalisation by $\mu_i(1-\mu_i)$ removes the dominant outer bound linking mean failure and variance, but it does not eliminate all finite-sample and geometric structure induced by the discreteness of the observed failure fractions and the small number of models. Accordingly, these correlations should be understood as exploratory descriptive summaries of this benchmark rather than as evidence of an independent structural compression effect. We include a summary of all calculated correlations in Table \ref{tab:length_correlations_comparison}.

\begin{table}[htb!]
\centering
\small
\begin{tabular}{lcc}
\hline
\textbf{Length Variable} & $\rho(\mu)$ & $\rho(\widetilde{\mathrm{Var}})$ \\
\hline
Prompt   & 0.28 & $-0.21$ \\
Solution & 0.32 & $-0.17$ \\
\hline
\end{tabular}
\caption{Task-level Spearman correlations between structural length variables and model behaviour. The second column reports the primary result of the paper, namely the association with mean failure. The third column reports a secondary exploratory association with difficulty-adjusted cross-model disagreement, computed on the 517 tasks with $0<\mu_i<1$.}
\label{tab:length_correlations_comparison}
\end{table}

\subsubsection{Hierarchical modelling of structural effects}

Our previous analyses so far reveal systematic relationships between structural length and model behavior. They do not, however, account for heterogeneity across model families. Because different LLMs exhibit distinct baseline failure rates and sensitivities to problem structure, we fit a hierarchical (mixed-effects) regression model to jointly capture global structural trends and model-specific deviations. We apply this framework separately to prompt length and solution length to determine whether the structural effects identified above persist when accounting for model-level variability.

We model the observed failure fraction $x_{i,m}$ for problem $i$ and model $m$ as a function of a log-transformed measure of length $L_{i}$,
\begin{equation} \label{eq: log_wordcount_L_i}
L_i = \log(1 + \text{word count}_i)\, .    
\end{equation}

The fitted model is
\begin{equation}\label{eq: hierarchical_model}
   x_{i,m} = \beta_0 +\beta_1 L_i + u_m +v_m L_i +\varepsilon_{i,m}, 
\end{equation}
where $\beta_0$ is the global intercept (measuring average baseline failure fraction), $\beta_1$ is the global average effect of structural length, $u_m$ is a model-specific random intercept which captures baseline performance differences between models, $v_m$ is a model-specific random slope for length that captures model-specific length sensitivity, and $\varepsilon_{i,m}$ is residual noise.
Throughout the analysis we assume $u_{m}$, $v_{m}$ and $\varepsilon_{i,m}$ to be normally distributed,

\begin{equation*}
\begin{pmatrix}
u_m \\
v_m
\end{pmatrix} \sim \mathcal{N}
\left(
\begin{pmatrix}
0 \\
0
\end{pmatrix},
\begin{pmatrix}
\sigma_u^2 & \sigma_{uv} \\
\sigma_{uv} & \sigma_v^2
\end{pmatrix}
\right)\, ,
\end{equation*}

\begin{equation*}
\qquad \varepsilon_{i,m} \sim \mathcal{N}(0, \sigma^2) \, .   
\end{equation*}

\subsubsection*{Prompt length}
\begin{figure*}[t]
\centering

\begin{subfigure}{0.48\linewidth}
\centering
\includegraphics[width=\linewidth]{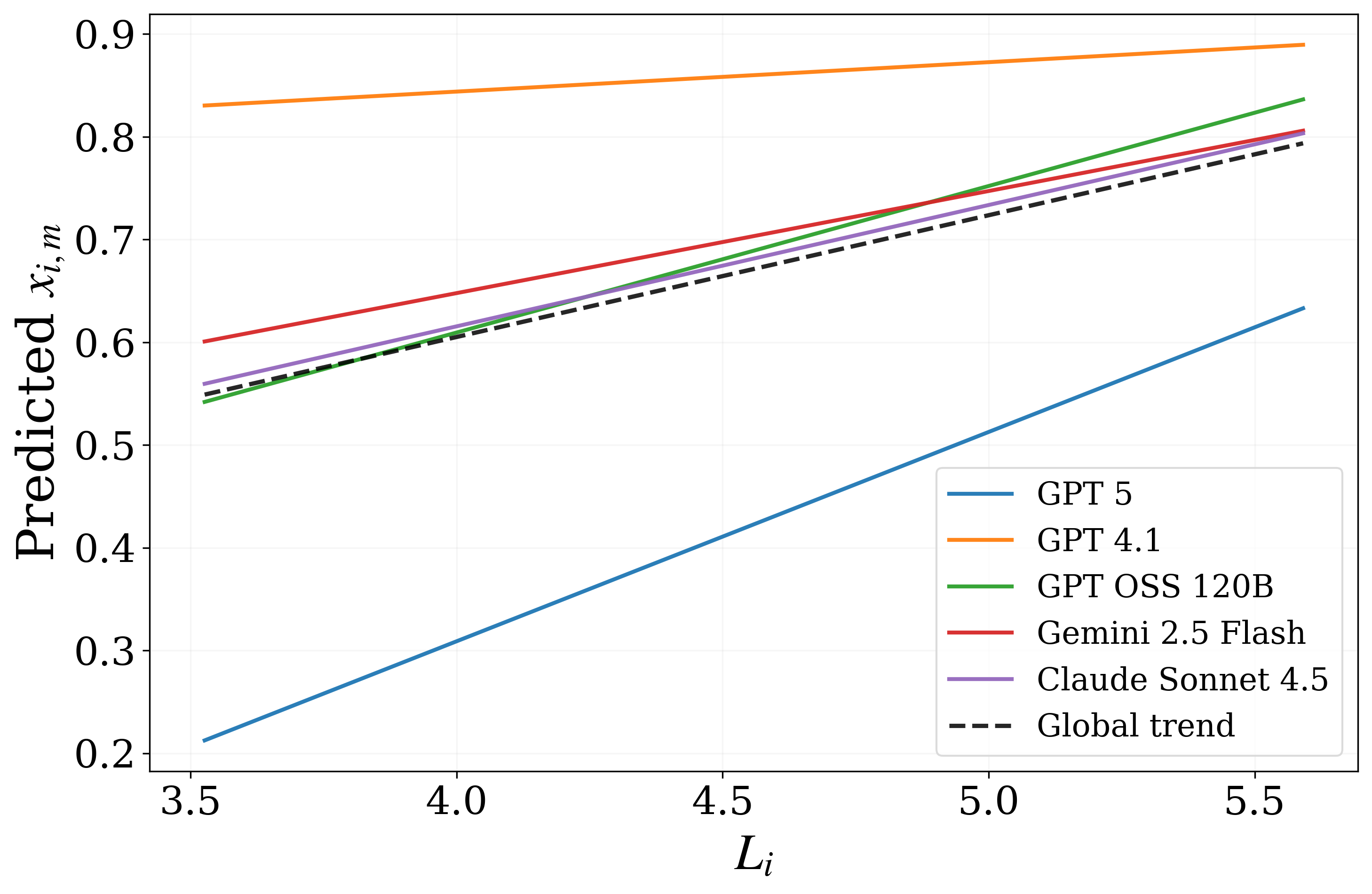}
\caption{Failure fraction as predicted by prompt length.}
\label{fig:hierarchical_fit_prompt_length}
\end{subfigure}
\hfill
\begin{subfigure}{0.48\linewidth}
\centering
\includegraphics[width=\linewidth]{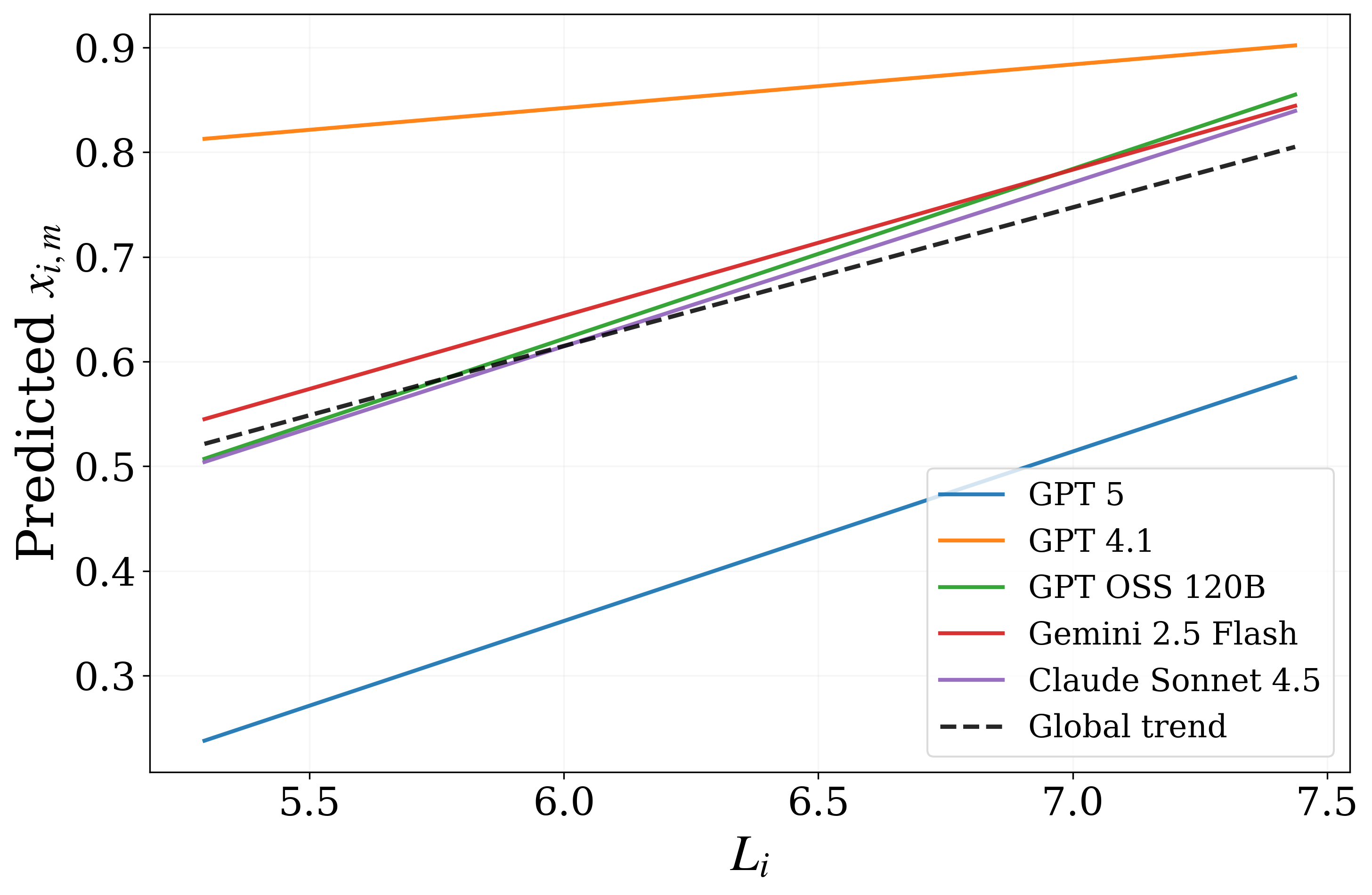}
\caption{Failure fraction as predicted by solution length.}
\label{fig:hierarchical_fit_sol_length}
\end{subfigure}

\caption{Predicted failure fraction $x_{i,m}$ as a function of $L_i = \log(1+\text{word count})$ under the fitted mixed-effects model. Solid lines show model-specific fits; the dashed line shows the global fixed effect.}
\label{fig:hierarchical_fit_both}

\end{figure*}

The model in \eqref{eq: hierarchical_model} was estimated for $L_{i}^{(\text{prompt})}$ via REML with $3035$ observations across the 5 LLMs under study. Table \ref{tab:hierarchical_table_prompt} presents a summary of these results, and figure \ref{fig:hierarchical_fit_prompt_length} represents the fitted hierarchical model by plotting the predicted failure trajectories for each LLM alongside the global fixed-effect trend.

\begin{table}[htb!]
\centering
\begin{tabular}{lccc}
\hline
\multicolumn{4}{c}{\textbf{Prompt Length Model}} \\
\hline
\textbf{Parameter} & \textbf{Estimate} & \textbf{SE} & \textbf{p-value} \\
\hline
$\beta_1$ & 0.118 & 0.037 &  0.001 \\
$\sigma_u^2$  & 0.281 & -- & -- \\
$\sigma_v^2$  & 0.006 & -- & -- \\
$\sigma_{uv}$  & $-0.042$ & -- & -- \\
$\sigma^2$  & 0.1153 & -- & -- \\
\hline
\end{tabular}
\caption{Mixed-effects model estimates for failure fraction $x_{i,m}$ as a function of log-transformed prompt length, $L_i^{(\text{prompt})}$.}
\label{tab:hierarchical_table_prompt}
\end{table}

\paragraph{Fixed effects} The estimated effect of $L_i^{(\text{prompt})}$ on failure fraction was $\beta_{1} = 0.118 \quad (SE = 0.037\, , p=0.001)$. This means a one-unit increase in $L_{i}^{(\text{prompt})}$ is associated with an average increase of approximately $0.118$ in $x_{i,m}$, which confirms the earlier Spearman correlation analysis, i.e. longer prompts are associated with higher failure rates. 

\paragraph{Random effects}

The random intercept variance, $\sigma_{u}^{2} = 0.281$, is substantial, indicating meaningful baseline differences in failure rates across models. This heterogeneity is consistent with the vertical separation between models observed in the binned scatterplots (Figure~\ref{fig:x_vs_promptlength}) and in the fitted trajectories shown in Figure~\ref{fig:hierarchical_fit_prompt_length}, and justifies the use of a hierarchical specification.

In contrast, the random slope variance, $\sigma_{v}^{2} = 0.006$, is very small, indicating minimal variation across models in sensitivity to prompt length. This is visually reflected in the near-parallel fitted lines in Figure~\ref{fig:hierarchical_fit_prompt_length}, suggesting that models degrade at comparable rates as prompt length increases.

Finally, the residual variance $\sigma^{2} = 0.1153$ remains substantial, indicating that additional problem-level factors beyond prompt length contribute to variation in $x_{i,m}$.

Overall, these results suggest that prompt length is associated with a broadly shared increase in failure across the evaluated models. Because the response variable is discrete and bounded, and because only five models are included, we interpret the random-slope estimates descriptively rather than as definitive evidence that all architectures respond identically to prompt length.

\subsubsection*{Solution length}

We estimated the same hierarchical model using $L_{i}^{(\text{solution})}$, the log-transformed word count of the reference solution. The model was fit using $3030$ observations across the five LLMs. Table~\ref{tab:hierarchical_table_sol} reports the parameter estimates, and Figure~\ref{fig:hierarchical_fit_sol_length} visualizes the corresponding fitted trajectories.

\begin{table}[htb!]
\centering
\begin{tabular}{lccc}
\hline
\multicolumn{4}{c}{\textbf{Solution Length Model}} \\
\hline
\textbf{Parameter} & \textbf{Estimate} & \textbf{SE} & \textbf{p-value} \\
\hline
$\beta_1$ & 0.137 & 0.026 & 0.001 \\
$\sigma_u^2$  & 0.225 & -- & -- \\
$\sigma_v^2$  & 0.003 & -- & -- \\
$\sigma_{uv}$  & -0.025 & -- & -- \\
$\sigma^2$  & 0.1128 & -- & -- \\
\hline
\end{tabular}
\caption{Mixed-effects model estimates for failure fraction $x_{i,m}$ as a function of log-transformed solution length, $L_i^{(\text{solution})}$.}
\label{tab:hierarchical_table_sol}
\end{table}

\paragraph{Fixed effects}

The fixed-effect coefficient for solution length is positive, $\beta_{1} = 0.137$, indicating that problems with longer reference solutions tend to produce higher failure fractions. This result is consistent with the earlier correlation analysis and suggests that tasks requiring longer solutions are generally more difficult for models to solve reliably.

\paragraph{Random effects}

As in the prompt-length model, the random intercept variance $\sigma_{u}^{2} = 0.225$ remains substantial, reflecting baseline differences in failure rates across models. However, variation in slopes across models is again limited ($\sigma_{v}^{2} = 0.003$), indicating that models respond similarly to increases in solution length.

The residual variance $\sigma^{2} = 0.1128$ remains non-negligible, indicating that solution length alone does not fully account for problem-level variation in failure rates.

Taken together, these results suggest that solution length is associated with increased task difficulty. Unlike prompt length, however, solution length is not directly observed by the model and likely reflects a mixture of intrinsic mathematical complexity, authoring style, and the granularity of the reference solution.

\section{Discussion}

Our work studies how two objectively measurable structural variables, prompt length and reference-solution length, relate to model failure on an adversarially constructed mathematics benchmark.

At the prompt level, we find that prompt length is a consistent predictor of empirical difficulty across all evaluated models: longer prompts are associated with higher mean failure rates. This relationship is also reflected in the mixed-effects analysis, whose fitted trends suggest that all five models degrade in performance as prompt length increases. At the same time, the fitted slopes are broadly similar, indicating that the models do not appear to differ dramatically in their sensitivity to prompt length at the level captured by this analysis.

At the level of problem-solution pairs, we likewise find that solution length is a significant predictor of model failure: longer reference solutions are associated with harder problems on average. This is consistent with prior work showing that mathematical reasoning difficulty often scales with the number of steps needed to arrive at a correct solution \citep{chain-of-thought}. However, the correlations of prompt length and solution length with model failure should each be interpreted differently. The reference solution is not shown to the model, so its length is best understood as a proxy for the amount of reasoning reflected in the human-written solution, rather than as an input-side feature of the prompt itself. On the other hand, prompt length emerges as a potential driver of model failure, meriting further exploration and ablation studies. 

The interpretation of cross-model disagreement is more delicate. Because disagreement measures based on variance are mechanically constrained by mean failure, raw variance correlations are difficult to interpret, especially in a dataset skewed toward harder tasks. For that reason, we treat the disagreement analysis as exploratory and focus on a simple normalised variance measure. Under this adjustment, both prompt length and solution length retain only weak negative associations with realised model separation. We therefore do not treat reduced discriminativeness as a headline result. At most, the evidence suggests that longer items may be somewhat less effective at separating models at a given difficulty level in this particular benchmark.

These findings have emerging relevance in the context of evaluation science, as they highlight a concrete way in which benchmark structure can affect evaluation outcomes. In mathematical settings, apparent model failures may reflect not only reasoning ability, but also structural properties of the task, such as the amount of information in the prompt or the complexity reflected by the reference solution. This matters in real-world settings where mathematical reasoning systems may be used for scientific, engineering, financial, or educational tasks: if evaluations do not account for structural confounders, they may overestimate or underestimate model reliability in specific deployment regimes.

Taken together, the clearest conclusion of the paper is that structural length is not innocuous: both prompt length and solution length are linked to empirical hardness in this benchmark. A secondary, more tentative conclusion is that structural length may also be related to realised model separation, although this part of the story is less clean because disagreement is geometrically and statistically constrained. For benchmark design, this reinforces the importance of analysing not only whether tasks are difficult, but also whether they remain informative for distinguishing model capabilities \citep{dynabench, leaderboard-illusion}. One practical implication is that mathematical benchmark reports should consider including length-controlled or length-stratified metrics, especially when comparing models on datasets with large variation in prompt or solution length.

A final limitation is that the reported correlations are necessarily contingent on the benchmark and model family under study. Our dataset was intentionally curated to be adversarial and difficult, and the analysis is based on a fixed set of five contemporary models evaluated under a specific repeated-attempt protocol. Accordingly, the quantitative relationships reported here should be understood as descriptive of this setting, rather than as universally stable estimates of how structural length affects difficulty or discriminativeness across all mathematical benchmarks and all model classes.

\section{Limitations and Future Work}

Several analyses in this work are best understood as descriptive first-pass summaries rather than final statistical treatments of the underlying phenomena. The response variable $x_{i,m}$ is discrete and bounded, since it is derived from only five attempts per model, and our disagreement measure is mechanically constrained by mean failure. Future work should therefore model the fail count $k_{i,m}$ directly, for example using a hierarchical binomial model with model-specific and problem-specific effects.

A richer model could also separate observed structural length from latent problem difficulty more explicitly. In the present analysis, prompt length and solution length are treated as measurable structural variables, but they may also correlate with unobserved mathematical complexity. Crossed random effects for problem and model, or latent difficulty parameters, would provide a more principled way to estimate how much variation is attributable to structural length rather than to unmeasured task difficulty.

A natural continuation of this work would be to extend the analysis to a broader set of models and benchmarks. Our model set was chosen to balance strong proprietary systems with an optimized open-weights model, but five models is still a small sample for drawing conclusions about model families. Replicating the analysis on additional reasoning systems, public benchmarks, and length-controlled task variants would help determine which effects are specific to this adversarial dataset and which are more general features of mathematical evaluation, from which more concrete proposals for improving model evaluations could be drawn.

Finally, word count is a coarse structural metric. More informative future analyses could examine notational density, number of stated conditions, number of reasoning branches, amount of irrelevant context, or the depth of the shortest valid solution. Such analyses would help clarify whether the observed length effects arise from surface verbosity, intrinsic mathematical complexity, or an interaction between the two.

\section*{Acknowledgments}

This work was supported by Poindexter Labs. We would like to thank every member of the team that participated in co-authoring this benchmark.

\bibliography{pdx}

\appendix
\section{Sample problem--solution pair}
\label{ap:samples}

To illustrate the style of the benchmark, we include one representative original problem--solution pair below.

\begin{sampleproblembox}
A carpenter wants a rectangle with positive integer side lengths $u$ and $v$ such that its area is
\[
uv=m^{19},
\]
and its side lengths satisfy the extra constraint
\[
v-3u\equiv 0 \pmod{14}
\]
for some $m$ in the set $\{1,2,\dots,30\}$.

For some values of $m$, no such rectangle exists. Find the sum of all such integers $m$.
\end{sampleproblembox}

\begin{samplesolutionbox}

\solstep{Step 1: Rewrite the congruence condition}

We seek integers $m\in\{1,2,\dots,30\}$ for which there do \emph{not} exist positive integers $u,v$ satisfying
\[
uv=m^{19}
\]
and
\[
v-3u\equiv 0 \pmod{14}.
\]

The latter congruence is equivalent to the system
\[
\begin{aligned}
v-3u &\equiv 0 \pmod{7},\\
v-3u &\equiv 0 \pmod{2}.
\end{aligned}
\]
Modulo $7$, this becomes
\[
v\equiv 3u \pmod{7},
\]
while modulo $2$, since $3\equiv 1\pmod{2}$, it becomes
\[
v\equiv u \pmod{2}.
\]

Thus we seek factor pairs $(u,v)$ of $m^{19}$ such that
\[
uv=m^{19},
\]
and also
\[
\begin{aligned}
v &\equiv 3u \pmod{7},\\
v &\equiv u \pmod{2}.
\end{aligned}
\]
Equivalently, $u$ and $v$ must have the same parity, and their residues modulo $7$ must satisfy $v\equiv 3u$.

\solstep{Step 2: Handle the case where $m$ is a multiple of $7$}

First consider $m$ divisible by $7$. In $\{1,2,\dots,30\}$, these are
\[
m\in\{7,14,21,28\}.
\]
Since $7\mid m$, we also have $7\mid m^{19}$. If we choose $u$ and $v$ both divisible by $7$, then automatically
\[
u\equiv v\equiv 0 \pmod{7},
\]
so
\[
v\equiv 3u \pmod{7}
\]
holds.

It remains only to match parity.

If $m$ is odd, namely $7$ or $21$, then $m^{19}$ is odd, so choosing
\[
u=7,
\qquad
v=\frac{m^{19}}{7}
\]
gives both $u$ and $v$ odd.

If $m$ is even, namely $14$ or $28$, then choosing
\[
u=14,
\qquad
v=\frac{m^{19}}{14}
\]
gives both $u$ and $v$ even.

Hence every multiple of $7$ in $\{1,\dots,30\}$ is admissible. Therefore none of
\[
7,14,21,28
\]
belongs to the set we seek.

\solstep{Step 3: Assume $7\nmid m$ and derive a necessary condition}

Now assume $7\nmid m$. Then $7\nmid m^{19}$, so any admissible $u$ and $v$ are invertible modulo $7$.

From
\[
uv=m^{19}
\]
and
\[
v\equiv 3u \pmod{7},
\]
multiplying the congruence by $v$ gives
\[
v^2\equiv 3uv\equiv 3m^{19}\pmod{7}.
\]
By Fermat's little theorem,
\[
m^6\equiv 1\pmod{7},
\]
hence
\[
m^{19}=m^{18}\cdot m\equiv m\pmod{7}.
\]
Therefore
\[
v^2\equiv 3m\pmod{7}.
\]

So a necessary condition for the existence of a valid rectangle is that $3m$ be a quadratic residue modulo $7$.

The quadratic residues modulo $7$ are
\[
0,\ 1,\ 2,\ 4.
\]
Since $7\nmid m$, we have $3m\not\equiv 0\pmod{7}$, so we need
\[
3m\equiv 1,\ 2,\ \text{or }4 \pmod{7}.
\]
The inverse of $3$ modulo $7$ is $5$, so this is equivalent to
\[
m\equiv 5,\ 3,\ \text{or }6 \pmod{7}.
\]
Thus if
\[
m\equiv 1,\ 2,\ \text{or }4 \pmod{7},
\]
then no valid rectangle can exist.

The numbers in $\{1,\dots,30\}$ in these residue classes are
\[
1,8,15,22,29,
\]
\[
2,9,16,23,30,
\]
\[
4,11,18,25.
\]
So the following $14$ values definitely fail:
\[
1,2,4,8,9,11,15,16,18,22,23,25,29,30.
\]

\solstep{Step 4: Examine the remaining residue classes}

We now examine the cases
\[
m\equiv 3,\ 5,\ \text{or }6 \pmod{7}.
\]

\medskip
\noindent\textbf{Case 4a:} $m\equiv 3\pmod{7}$.

These are
\[
3,10,17,24.
\]
Here
\[
3m\equiv 2\pmod{7},
\]
so we need
\[
v^2\equiv 2\pmod{7},
\]
whose solutions are
\[
v\equiv 3,\ 4 \pmod{7}.
\]

For each of these values, taking $v=m$ works:
\[
\begin{aligned}
3&\equiv 3, & 10&\equiv 3,\\
17&\equiv 3, & 24&\equiv 3
\pmod{7}.
\end{aligned}
\]
Also $u=m^{18}$ has the same parity as $v=m$. Hence all four values are admissible.

\medskip
\noindent\textbf{Case 4b:} $m\equiv 5\pmod{7}$.

These are
\[
5,12,19,26.
\]
Now
\[
3m\equiv 1\pmod{7},
\]
so we need
\[
v^2\equiv 1\pmod{7},
\]
whose solutions are
\[
v\equiv 1,\ 6 \pmod{7}.
\]

Taking $v=m^3$ works in each case, because
\[
5^3\equiv 6\pmod{7},
\]
and hence similarly
\[
12^3\equiv 19^3\equiv 26^3\equiv 6\pmod{7}.
\]
Then $u=m^{16}$, so $u$ and $v$ have the same parity. Thus all four values are admissible.

\medskip
\noindent\textbf{Case 4c:} $m\equiv 6\pmod{7}$.

These are
\[
6,13,20,27.
\]
Now
\[
3m\equiv 4\pmod{7},
\]
so we need
\[
v^2\equiv 4\pmod{7},
\]
whose solutions are
\[
v\equiv 2,\ 5 \pmod{7}.
\]

For $m=6$, the divisors of $6^{19}$ include $2$, and
\[
2^2\equiv 4\pmod{7}.
\]
Taking $v=2$ works, and then
\[
u=\frac{6^{19}}{2}
\]
is even, so parity also matches.

For $m=20$, the same choice $v=2$ works, since $2\mid 20^{19}$.

For $m=27$, the divisors are powers of $3$, and taking
\[
v=3^2=9
\]
gives
\[
v\equiv 2\pmod{7},
\qquad
v^2\equiv 4\pmod{7}.
\]
Also both $v$ and
\[
u=\frac{27^{19}}{9}=3^{55}
\]
are odd, so parity matches.

For $m=13$, however, every divisor is a power of $13$, and
\[
13\equiv -1\pmod{7}.
\]
So every divisor is congruent to either $1$ or $6$ modulo $7$, and squaring either gives
\[
1^2\equiv 1,\qquad
6^2\equiv 1
\pmod{7}.
\]
Thus no divisor $v$ of $13^{19}$ can satisfy
\[
v^2\equiv 4\pmod{7}.
\]
Therefore $m=13$ is not admissible.

\solstep{Step 5: Collect the failing values and sum them}

The full set of $m\in\{1,\dots,30\}$ for which no such rectangle exists is
\[
\{1,2,4,8,9,11,13,15,16,18,22,23,25,29,30\}.
\]
Their sum is
\begin{align*}
&1+2+4+8+9+11+13+15 \\
&+16+18+22+23+25+29+30 \\
&= (1+29)+(2+30)+(4+25)+(8+22) \\
&\quad +(9+23)+(11+18)+(13+16)+15 \\
&= 30+32+29+30+32+29+29+15 \\
&= 226.
\end{align*}

Therefore the answer is
\[
\boxed{226}.
\]

\end{samplesolutionbox}

\end{document}